\pdfoutput=1
\PassOptionsToPackage{hyphens}{url}
\PassOptionsToPackage{breaklinks}{hyperref}

\documentclass[letterpaper]{article}
\usepackage{ijcai20}

\usepackage{times}
\usepackage{soul}
\usepackage{hyperref}
\usepackage{url}
\usepackage[utf8]{inputenc}
\usepackage[small]{caption}
\usepackage{graphicx}
\usepackage{amsmath}
\usepackage{amsthm}
\usepackage{booktabs}
\usepackage{algorithm}
\usepackage{algorithmic}
\usepackage{xcolor}

\newtheorem{example}{Example}

\newtheorem{definition}{Definition} 

\usepackage{todonotes}
\usepackage{subcaption}
\usepackage{tikz, pgfplots}
\usetikzlibrary{automata, arrows, shapes}

\newcommand{\cf}{\textsf{cf}}
\newcommand{\adm}{\textsf{adm}}
\newcommand{\com}{\textsf{com}}
\newcommand{\prf}{\textsf{prf}}
\newcommand{\grd}{\textsf{grd}}
\newcommand{\stb}{\textsf{stb}}

\newcommand{\enum}{\mathsf{Enum}}
\newcommand{\scept}{\mathsf{Scept}}
\newcommand{\cred}{\mathsf{Cred}}
\newcommand{\constr}{\mathsf{Constr}}

\title{Deep Learning for Abstract Argumentation Semantics}

\author{
Dennis Craandijk$^{1,2}$
\And
Floris Bex$^{2,3}$
\affiliations
$^1$ National Police Lab AI, Netherlands Police\\
$^2$ Information and Computing Sciences, Utrecht University \\
$^3$ Institute for Law, Technology and Society, Tilburg University\\
\emails
\{d.f.w.craandijk, f.j.bex\}@uu.nl
}

\begin{document}

\maketitle

\begin{abstract}
In this paper, we present a learning-based approach to determining acceptance of arguments under several abstract argumentation semantics. More specifically, we propose an argumentation graph neural network (AGNN) that learns a message-passing algorithm to predict the likelihood of an argument being accepted. The experimental results demonstrate that the AGNN can almost perfectly predict the acceptability under different semantics and scales well for larger argumentation frameworks. Furthermore, analysing the behaviour of the message-passing algorithm shows that the AGNN learns to adhere to basic principles of argument semantics as identified in the literature, and can thus be trained to predict extensions under the different semantics – we show how the latter can be done for multi-extension semantics by using AGNNs to guide a basic search. We publish our code at \url{https://github.com/DennisCraandijk/DL-Abstract-Argumentation}. 
\end{abstract}

\section{Introduction}
Over the past few years an increasing amount of research effort has been directed towards designing deep learning models that learn on problems from symbolic domains \cite{garcez2015neural}. Recent progress has sparked interest in graph neural networks (GNNs), a class of neural networks capable of performing computations over graphs. Due to their strong relational inductive bias \cite{battaglia2018relational}, GNNs can be trained to solve constraint satisfaction problems which require performing relational inferences such as boolean satisfiability \cite{selsam2018learning} and solving Sudoku puzzles \cite{palm2018recurrent}.

One domain in symbolic AI that is relatively unexplored with respect to GNNs is \emph{computational argumentation}, an approach to defeasible reasoning that focuses on interactions between arguments and counterarguments. With applications in multi-agent systems, decision-making tools, medical and legal-reasoning, argumentation has become a major subfield of AI~\cite{atkinson2017towards}. Much of the theory in computational argumentation is built on Dung's~\shortcite{dung1995acceptability} pioneering work on abstract argumentation frameworks, which introduced several acceptability semantics that define which sets of arguments (\emph{extensions}) can be reasonably accepted given an argumentation framework (AF) of arguments and attacks between these arguments, often represented as a graph. Thus, it can be determined if an argument can be accepted given an AF by looking at whether it is contained in some extensions (credulous acceptance) or all extensions (sceptical acceptance) under a given semantics. Due to the computational complexity of determining which arguments can be accepted, the design of efficient methods for computing extensions and acceptability constitutes an active research direction within the argumentation community. Most current approaches solve acceptance problems by translating the problem to a symbolic formalism for which a dedicated solver exists, such as constraint-satisfaction problems, propositional logic or answer-set programming~\cite{gaggl2020design,charwat2015methods}.

In this paper, we propose an argumentation graph neural network (AGNN) that learns to predict credulous and sceptical acceptance of arguments under 4 well-known argumentation semantics. To the best of our knowledge, only Kulhman and Thimm~\shortcite{kuhlmann2019using} have conducted a preliminary study using GNNs, implementing a conventional single forward pass classifier to approximate credulous acceptance under the preferred semantics with an average class accuracy of around 0.61. We propose a recurrent GNN that can almost perfectly predict (MCC between 0.997 and 1) both credulous and sceptical acceptance under several semantics by learning to perform a sequence of relational inferences based on the attack relations between the arguments in an AF. Furthermore, we also provide a way to predict (multiple) extensions given an AF by using AGNN to guide a search procedure. 

Our learning-based approach to determining argument acceptance shows that sub-symbolic deep learning techniques can accurately solve a problem that could previously only be solved by sophisticated symbolic solvers. By inspecting the behaviour of the message-passing algorithm of AGNN, we see that it has learnt some basic principles of argumentation semantics \cite{baroni2011introduction}, and in the case of acceptance under the grounded semantics exhibits behaviour similar to a well-established symbolic labelling algorithm for the grounded extension \cite{modgil2009proof}. AGNN is a single architecture that is able to approximate argumentation problems of differing complexity and of sizes substantially larger than what it saw during training in constant time, simply by running for more iterations. While symbolic solvers always provide the correct answer, different problems under different semantics each need their own tailor-made algorithm, and the time complexity depends on the complexity of the problem.

\section{Preliminaries}
\label{sec:preliminaries}
We recall Dung's abstract argumentation frameworks~\shortcite{dung1995acceptability}.

\begin{definition}
\label{def:AF}
An abstract \emph{argumentation framework} (AF) is a pair $F = (A, R)$ where A is a (finite) set of arguments and $R \subseteq A \times A$ is the attack relation. The pair $(a,b) \in R$ means that $a$ \emph{attacks} $b$. A set $S \subseteq A$ attacks $b$ if there is an $a \in S$, such that $(a,b) \in R$. An argument $a \in A$ is \emph{defended} by $S \subseteq A$ iff, for each $b \in A$ such that $(b,a) \in R$, $S$ attacks $b$.
\end{definition}

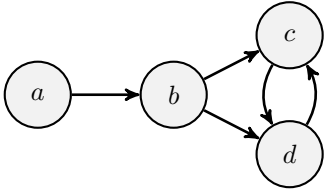
\begin{figure}
	\centering
    \begin{tikzpicture}[
        ->, 
    	>=stealth', 
    	every state/.style={thick, fill=gray!10}, 
    	every edge/.append style={line width=1pt},
    	node distance=0.15cm and 0.9cm
    ]
    \node[state] (a) {$a$};
    \node[state, right= of a] (b) {$b$};
    \node[state, above right = of b] (c) {$c$};
    \node[state, below right = of b] (d) {$d$};
    \draw   (a) edge (b)
            (b) edge (c)
            (b) edge (d)
            (c) edge [bend right] (d)
            (d) edge [bend right] (c)
            ;
    \end{tikzpicture}
    \caption{Graph representations of the AF $F_{e}$.}
    \label{fig:AF}
\end{figure}

\begin{example}
Figure \ref{fig:AF} illustrates the AF $F_{e} = (\{a,b,c,d\},\allowbreak \{(a,b), (b,c), (b,d), (c,d), (d,c)\})$, which serves as a running example.
\label{example:AF}
\end{example}

Dung-style semantics define the sets of arguments that can jointly be accepted (\emph{extensions}). A $\sigma$-extension refers to an extension under semantics $\sigma$. We consider admissible sets and preferred, complete, grounded and stable semantics with the following functions respectively \adm, \prf, \com, \grd, \stb.

\begin{definition}
Let $F=(A,R)$ be an AF. A set $S \subseteq A$ is \emph{conflict-free} (in F), if there are no $a,b \in S$, such that $(a,b) \in R$. The collection of sets which are conflict-free is denoted by $\cf(F)$. For $S \in \cf(F)$, it holds that:

\begin{itemize}
     \item $S \in \adm(F)$, if each $a \in S$ is defended by $S$;
     \item $S \in \prf(F)$, if $S \in \adm(F)$ and for each $T \in \adm(F)$, $S \not\subset T$;
     \item $S \in \com(F)$, if $S \in \adm(F)$ and for each $a \in A$ defended by $S$ it holds that $a \in S$;
     \item $S \in \grd(F)$, if $S \in \com(F)$ and for each $T \in \com(F)$, $T \not\subset S$;
     \item $S \in \stb(F)$, if for each $a \in A \setminus S$, S attacks $a$.
 \end{itemize}
 
\end{definition}

\begin{example}
The extensions of $F_{e}$ under the preferred, complete and grounded semantics are: $\prf(F) = \{\{a,c\}, \{a,d\}\}; \allowbreak \com(F) = \{\{a\}, \{a,c\}, \{a,d\}\}; \allowbreak \grd(F) = \{a\}; \allowbreak \stb(F) = \{\{a,c\}, \{a,d\}\}$.
\label{example:semantics}
\end{example}

Typical problems of interest for abstract argumentation semantics are as follows.
\begin{definition}
Given an AF $F=(A,R)$, a semantics $\sigma$ and some argument $a \in A$:
\begin{itemize}
    \item \emph{Enumeration} $\enum_{\sigma}$: construct all extensions prescribed by $\sigma$
    \item \emph{Credulous acceptance} $\cred_{\sigma}$: decide if $a$ is contained in at least one $\sigma$-extension
    \item \emph{Sceptical acceptance} $\scept_{\sigma}$: decide if $a$ is contained in all $\sigma$-extensions
\end{itemize}
\end{definition}

\begin{example}
Under the preferred semantics, only argument $a$ is sceptically accepted and arguments $a$, $c$ and $d$ are credulously accepted in $F_{e}$.
\end{example}

The problem of deciding the acceptability of arguments is well-studied \cite{dunne2009complexity}. Whereas $\cred_{\grd}$, $\scept_{\grd}$ and $\scept_{\com}$ can be solved in polynomial time, all other problems considered here are shown to be to NP-complete or surpassing.

\section{Problem Setup}\label{sec:problem}

In order to solve argument acceptability problems with a GNN we pose them as a classification problem. Consider an AF $F=(A,R)$ and a semantics $\sigma$. Our goal is to approximate a function $f_{\sigma}$ mapping the input $F$ to a binary labelling $f_{\sigma}(F)$ denoting the acceptability of all arguments in $A$ under semantics $\sigma$. The function is approximated by producing a value for each argument in the interval $[0,1]$ - representing the likelihood whether an argument can be accepted - which is rounded to produce a binary answer (accept or reject).

\section{Model}\label{sec:model}
We introduce our \emph{argumentation graph neural network} (AGNN) model. An AGNN maps an AF to a graph representation and assigns a multidimensional embedding to each node. These embeddings are then iteratively updated by performing a number of message passing steps. At each iteration nodes broadcast their embeddings by exchanging messages with their neighbours and subsequently update their embedding based on the incoming messages. After each iteration those embeddings can be read out to produce the predicted likelihood of the respective argument being accepted.

More formally, $G$ is an AF graph representation in which arguments are nodes and attacks are directed edges. Each node $i$ is assigned an embedding, denoted by $v^{t}_{i}$\ at step $t$. The node embedding is initialised by a learned embedding $x_{i}$ such that $v^{0}_{i} = x_{i}$. Each message passing iteration the embeddings are updated according to:

\begin{align}
\begin{split}
m^{t + 1}_{i} = & \sum_{j \in N^{s}(i)} M^{s}(v^{t}_{i}, v^{t}_{j}) + \\ & \sum_{k \in N^{t}(i)} M^{t}(v^{t}_{i}, v^{t}_{k}) 
\end{split}
\\
(v^{t+1}_{i}, h^{t+1}_{i}) = & U(h^{t}_{i}, m^{t + 1}_{i}, x_{i})
\end{align}

where $N^{s}(i)$ and $N^{t}(i)$ denote all nodes which have a connection with node $i$ and for which $i$ is the source or target node respectively. The message functions $M^{s}$ and $M^{t}$ are Multilayer perceptrons (MLPs) which learn to compute a message to send along edges based on the embeddings of the nodes it connects. $M^{s}$ computes a message from the source node to the target node and $M^{t}$ vice versa. Messages from all neighbours are subsequently summed to form the incoming message $m^{t}_{i}$. Aggregating messages into a single incoming message allows the model to handle graphs of arbitrary size. For nodes which do not have any incoming edges $m^{t}_{i}$ is filled with zeros. The update function $U$ is a Recurrent Neural Network (RNN) which learns how to update a node given the incoming message and the node's input feature, where $h^{t}_{i}$ is the RNNs hidden state. By updating the node embeddings recurrently while also accounting for the input features, AGNN is able to iteratively refine embeddings without forgetting any potentially relevant information. 

After each iteration the embeddings can be read out with the readout function $R$. $R$ is an MLP that learns to map a node's embedding $v^{t}_{i}$ to a logit probability $o^t_{v_{i}} = R(v^{t}_{i})$ representing the likelihood of the respective argument being accepted. This logit probability can subsequently be converted to a likelihood in the interval $[0,1]$ using a sigmoid function. 

The message and update functions form the core of the AGNN model. Together, the functions yield a neural message passing algorithm whose parameters can be optimised. The readout function serves as a mapping between the multidimensional embeddings and the output values. In terms of argumentation AGNN learns how to initialise arguments with an embedding; recurrently update these embeddings by exchanging messages between arguments over the attack relations; and map the argument embeddings to a likelihood of that argument being accepted. 

\begin{table}
\centering
\setlength\tabcolsep{3pt} 
\begin{tabular}{lrrrr} 

\toprule
Characteristic & \grd & \prf & \stb & \com \\ 
\midrule
Extensions per AF & 1.0 & 2.1 & 1.6 & 6.3 \\
Arguments per extension & 4.7 & 9.5 & 11.8 & 8.0 \\
Scept. accepted arguments per AF & 4.8 & 5.9 & 5.8 & 4.8 \\
Cred. accepted arguments per AF & 4.8 & 8.0 & 7.9 & 8.0 \\
\bottomrule
\end{tabular}
\caption{AF characteristics averaged over all AFs the test dataset.}
\label{tab:data}
\end{table}

\section{Experimental Setup}\label{sec:setup}
\subsection{Data}
\label{sec:data}
We generate a variety of challenging argumentation frameworks by sampling from the following AF generators from the International Competition on Computational Models of Argumentation \cite{gaggl2020design}: \emph{AFBenchGen2}, \emph{AFGen Benchmark Generator}, \emph{GroundedGenerator}, \emph{SccGenerator}, \emph{StableGenerator}. To avoid duplicates, each AF is checked for isomorphism with \emph{Nauty}~\cite{mckay2014practical}. Ground-truth labels are determined based on extensions obtained with the sound and complete \emph{$\mu$-toksia} solver~\cite{niskanen2019mu}. We generate a test and validation dataset of size 1000 with AFs containing $|A|=25$ arguments, and a training dataset of a million AFs where the number of arguments per AF is sampled randomly between $5\leq|A|\leq25$ (to accelerate the learning). Table~\ref{tab:data} shows characteristics of the AFs in the test dataset under different semantics.

\subsection{Training}
\label{sec:training}
We instantiate the AGNN model with one hidden layer and a rectified linear unit for non-linearity for the MLPs $M^{s}$, $M^{t}$ and $R$, a Long Short-Term Memory \cite{hochreiter1997lstm} for $U$ and a shared random embedding $x_{i}$ for all nodes. The dimensions of the embedding and all hidden neural layers are $d=128$. The model is run for $\mathcal{T}=32$ message passing steps. We train our model in batches containing 50 graphs (approximately 750 nodes) using the AdamW optimiser \cite{loshchilov2017decoupled} with a cosine cyclical learning rate \cite{smith2017cyclical} between $2\text{e}^{-4}$ and $1\text{e}^{-7}$, $\ell_{2}$ regularisation of $1\text{e}^{-9}$ and clip the gradients by global norm with a $0.5$ clipping ratio~\cite{pascanu2013difficulty}. We train the model by minimising the binary cross entropy loss between the predicted likelihood and the ground-truth binary label. We minimise the loss at every message passing step (rather than only on the final step) since this encourages the model to learn a convergent message passing algorithm while also mitigating the vanishing gradient problem~\cite{palm2018recurrent}.

\begin{table}
\centering
\setlength\tabcolsep{2pt} 
\resizebox{\columnwidth}{!}{%
\begin{tabular}{llrrrrlrrrr} 

\toprule
 & & \multicolumn{4}{c}{$\scept_{\sigma}$} && \multicolumn{4}{c}{$\cred_{\sigma}$} \\ 
 \cmidrule{3-6} \cmidrule{8-11} 
Metric & Model & \grd & \prf & \stb & \com && \grd & \prf & \stb & \com \\ 
\midrule 
& GCN & 0.17 & 0.18 & 0.20 & 0.16 && 0.17 & 0.17 & 0.39 & 0.36 \\
MCC & FM2 & 0.64 & 0.54 & 0.55 & 0.64 && 0.63 & 0.57 & 0.55 & 0.57 \\
& AGNN & \textbf{1.00} & \textbf{0.997} & \textbf{0.997} & \textbf{1.00} && \textbf{1.00} & \textbf{0.998} & \textbf{0.998} & \textbf{0.999} \\
\midrule
MAE & AGNN & 3$\text{e}^{-8}$ & 5$\text{e}^{-4}$ & 9$\text{e}^{-4}$ & 3$\text{e}^{-8}$ && 3$\text{e}^{-8}$ & 6$\text{e}^{-4}$ & 4$\text{e}^{-4}$ & 3$\text{e}^{-4}$ \\

\bottomrule
\end{tabular}}
\caption{Argument acceptance results on the test dataset.}
\label{tab:result}
\end{table}

\section{Results}\label{sec:experiments}
We train AGGN on the AFs in the training dataset for each acceptance problem and semantics described in Section~\ref{sec:preliminaries}. We use the Matthews Correlation Coefficient (MCC) to evaluate the binary classification performance on the AFs in the test dataset. Since accepted and rejected arguments are equally important but unequally distributed in the dataset (see Table~\ref{tab:data}), MCC provides a more balanced metric compared to accuracy or the F1-score~\cite{powers2011evaluation}. We compare our approach to a graph convolutional network (GCN)~\cite{kipf2017semi} baseline and our own implementation of the FM2 model of Kuhlmann and Thimm \shortcite{kuhlmann2019using}. Both are single forward pass classifiers, where FM2 is a GCN with the number of incoming and outgoing attacks per argument added as input features. Table~\ref{tab:result} reports the MCC scores for all models. AGNN performs considerably better than GCN or FM2, and achieves a perfect score on all problems which belong to complexity class P. On all other problems (belonging to NP or surpassing) AGNN is able to correctly predict the acceptance of arguments almost perfectly.

In addition to the classification performance of the model, we are interested in the confidence of its predictions. As a measure of prediction confidence we use the mean absolute error (MAE) between the predicted likelihoods and the binary ground-truth labels. An MAE of $0.01$ implies that on average the predicted likelihood deviates 1 percentage point from the ground-truth label. A low MAE thus indicates predictions are made correctly and with high confidence. Table~\ref{tab:result} shows that predictions are overall made with high confidence. Most notably AGNN predictions deviate only $3e^{-6}$ percentage points from the true label on all problems for which it achieved a perfect classification score.

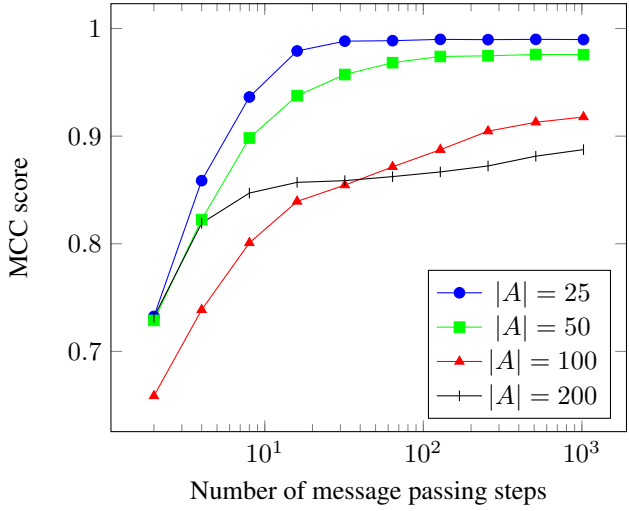
\begin{figure}
	\centering
	\begin{tikzpicture}
	\begin{axis}[
    xlabel={Number of message passing steps},
    ylabel={MCC score},
    xmode=log,
    legend pos=south east,
]
 
\addplot[
    color=blue,
    mark=*,
    ]
    coordinates {(2, 0.7324400168782608)(4, 0.8586381289496985)(8, 0.9363635970321076)(16, 0.9791517592451568)(32, 0.9882664612952438)(64, 0.988705092179646)(128, 0.9899325262076548)(256, 0.989670806639488)(512, 0.989933620880504)(1024, 0.9897585501526908)
    };
\addplot[
    color=green,
    mark=square*,
    ]   
    coordinates {(2, 0.7287428249074855)(4, 0.8224959616747973)(8, 0.8982625474614172)(16, 0.9375775408442896)(32, 0.9572500956155482)(64, 0.968263906493238)(128, 0.9740127894160342)(256, 0.9746411003450164)(512, 0.9758745103794972)(1024, 0.9757348472197772)
    };
\addplot[
    color=red,
    mark=triangle*,
    ]
	coordinates {(2, 0.6585293965844752)(4, 0.7385917612538105)(8, 0.8008103238378431)(16, 0.8393304837076969)(32, 0.8545174254402919)(64, 0.8714929749320359)(128, 0.8873734770912485)(256, 0.90462688915182)(512, 0.9128910477087544)(1024, 0.9177802745689676)
    };
    
\addplot[
    color=black,
    mark=|,
    ]
	coordinates {(2, 0.7317802664436499)(4, 0.8191290796239007)(8, 0.8471200367078193)(16, 0.8570466475339239)(32, 0.8586335231848754)(64, 0.8624014332838681)(128, 0.8667298008418365)(256, 0.8722851131505491)(512, 0.88139001553449)(1024, 0.8874846025254567)
    };
    \legend{$|A|=25$, $|A|=50$, $|A|=100$, $|A|=200$}
\end{axis}
	\end{tikzpicture}
	\caption{MCC score for $\cred_{\prf}$ on AFs of different size as a function of the number of message passing steps $\mathcal{T}$. The performance for the first few message passing steps mainly reflects how well AGNN is able to anticipate the status of arguments before convergence. Because AFs are randomly generated the performance during this anticipation phase may vary between datasets, hence the crossing lines for $|A|=100$ and $|A|=200$.}
	\label{fig:scale}
\end{figure}

\subsection{Scaling}
Even though AGNN is trained on AFs of size $5\leq|A|\leq25$, it is able to determine acceptability in much larger AFs. In order to test how well the trained model scales to larger instances we generate extra test datasets for each $|A| \in \{50,100,200\}$ with 1000 AFs containing $|A|$ arguments. Figure~\ref{fig:scale} illustrates the MCC scores for predicting the credulous acceptance under the preferred semantics on different sized AFs as a function of the number of message passing steps $\mathcal{T}$. The figure shows AGNN continues to improve its predictions on large AFs by running for more iterations. Notably the performance on $|A|=200$ AFs still improves after hundreds of iterations (which is not surprising considering that those AFs on average contain 5e\textsuperscript{3} attacks and up to 9e\textsuperscript{5} extensions). This indicates that, while only being trained to perform 32 message passing steps, AGNN has learned some general and convergent message passing procedure which scales to larger AFs by performing more iterations. 

AGNN exhibits similar behaviour on all problems of the same complexity as $\cred_{\prf}$. On all problems belonging to complexity class P, AGNN is able to correctly classify all arguments in AFs with $|A|=200$ when run for 32 message passing iterations.

\section{Analysing AGNN Behaviour}
\label{sec:message_passing}
The AGNN model learns a message passing algorithm which enables it to predict the acceptance status of arguments in an AF. The process of iteratively updating arguments by exchanging messages can be understood as performing a sequence of relational inferences between connected arguments. Since the computations underlying those inferences are learned by neural networks, it is hard to interpret `how' those inferences enable the model to predict acceptance.

We inspect the outputs of each iteration in order to infer how the model works towards a solution. AGNN exhibits similar behaviour on all AFs in the test dataset. Arguments are initialised with a low confidence prediction. At each iteration the likelihoods change based on the incoming messages, until arguments `decide' on their status by \emph{converging} to a high confidence likelihood close to $0$ or $1$. Generally, the convergence of an argument directly affects the prediction of adjacent arguments in the next iteration. As information is exchanged between arguments, convergence propagates through the graph until the model stops evolving and the likelihoods stay more or less constant.

\begin{figure}
\centering
\setlength\tabcolsep{4pt} 
\resizebox{\columnwidth}{!}{%
\begin{tabular}{cccc} 
\toprule
 &\multicolumn{3}{c}{$t$} \\ 
 \cmidrule{2-4}
 AF $F$ & 1 & 2 & 3 \\ 
\midrule

\begin{tikzpicture}[
    ->, 
    >=stealth', 
    every node/.style={fill=gray!10},
    every edge/.append style={line width=0.6pt},
    node distance=1.1cm and 0.25cm]

  \node (0) [draw, state] {a};
  \node (1) [draw,state, below left = of 0] {b};
  \node (3) [draw,state, below right = of 0] {c};
  \node (2)  [draw,state, below right = of 1] {d};
  \draw [->] (0) edge (1);
  \draw [->] (0)edge (3);
  \draw [->] (1) edge [bend right](3);
  \draw [->] (1) edge(2);
  \draw [->] (3) edge [bend right](1);
  \draw [->] (2) edge (3);
\end{tikzpicture}

&

\begin{tikzpicture}[
    ->, 
    >=stealth', 
    every node/.style={fill=gray!10},
    every edge/.append style={line width=0.6pt},
    node distance=1.1cm and 0.25cm]
\definecolor{fillcolor}{rgb}{0.18,0.18,1.0};
  \node (0) [draw,fill=fillcolor, state] {\textsf{A}};
  \definecolor{fillcolor}{rgb}{1.0,0.92,0.92};
  \node (1) [draw,fill=fillcolor,state, below left = of 0] {\textsf{R}};
  \definecolor{fillcolor}{rgb}{1.0,0.89,0.89};
  \node (3) [draw,fill=fillcolor,state, below right = of 0] {\textsf{R}};
  \definecolor{fillcolor}{rgb}{0.99,0.99,1.0};
  \node (2)  [draw,fill=fillcolor,state, below right = of 1] {\textsf{R}};
  \draw [->] (0) edge (1);
  \draw [->] (0)edge (3);
  \draw [->] (1) edge [bend right](3);
  \draw [->] (1) edge(2);
  \draw [->] (3) edge [bend right](1);
  \draw [->] (2) edge (3);
\end{tikzpicture}

&

\begin{tikzpicture}[
    ->, 
    >=stealth', 
    every node/.style={fill=gray!10},
    every edge/.append style={line width=0.6pt},
    node distance=1.1cm and 0.25cm]
\definecolor{fillcolor}{rgb}{0.22,0.22,1.0};
  \node (0) [draw,fill=fillcolor, state] {\textsf{A}};
  \definecolor{fillcolor}{rgb}{1.0,0.41,0.41};
  \node (1) [draw,fill=fillcolor,state, below left = of 0] {\textsf{R}};
  \definecolor{fillcolor}{rgb}{1.0,0.3,0.3};
  \node (3) [draw,fill=fillcolor,state, below right = of 0] {\textsf{R}};
  \definecolor{fillcolor}{rgb}{0.93,0.93,1.0};
  \node (2)  [draw,fill=fillcolor,state, below right = of 1] {\textsf{A}};
  \draw [->] (0) edge (1);
  \draw [->] (0)edge (3);
  \draw [->] (1) edge [bend right](3);
  \draw [->] (1) edge(2);
  \draw [->] (3) edge [bend right](1);
  \draw [->] (2) edge (3);
\end{tikzpicture}

& \begin{tikzpicture}[
    ->, 
    >=stealth', 
    every node/.style={fill=gray!10},
    every edge/.append style={line width=0.6pt},
    node distance=1.1cm and 0.25cm]
\definecolor{fillcolor}{rgb}{0.0,0.0,1.0};
  \node (0) [draw,fill=fillcolor, state] {\textsf{A}};
  \definecolor{fillcolor}{rgb}{1.0,0.24,0.24};
  \node (1) [draw,fill=fillcolor,state, below left = of 0] {\textsf{R}};
  \definecolor{fillcolor}{rgb}{1.0,0.28,0.28};
  \node (3) [draw,fill=fillcolor,state, below right = of 0] {\textsf{R}};
  \definecolor{fillcolor}{rgb}{0.2,0.2,1.0};
  \node (2)  [draw,fill=fillcolor,state, below right = of 1] {\textsf{A}};
  \draw [->] (0) edge (1);
  \draw [->] (0)edge (3);
  \draw [->] (1) edge [bend right](3);
  \draw [->] (1) edge(2);
  \draw [->] (3) edge [bend right](1);
  \draw [->] (2) edge (3);
\end{tikzpicture}    \\
\bottomrule
\end{tabular}}
\caption{The acceptance predictions AGNN makes after the first three message passing iterations on the AF $F=(\{a,b,c,d\},\{(a,b),(a,c),(b,c),(b,d),(c,b),(d,c)\}$ with respect to the grounded semantics. The label and colour of each arguments denote whether the argument is predicted to be \textsf{\textsf{A}} accepted or \textsf{\textsf{R}} rejected where a darker colour indicates a higher confidence prediction. At $t=1$ argument $a$ converges to \emph{accept} while the other arguments anticipate \emph{reject} with a confidence that positively correlates with the amount of incoming attacks. At $t=2$ arguments $b$ and $c$ converge to \emph{reject} while $d$ adjusts its anticipation to \emph{accept} since all its neighbours anticipated \emph{reject}. At $t=3$ $d$ converges to \emph{accept} after which the model stops evolving.}
\label{fig:message}
\end{figure}
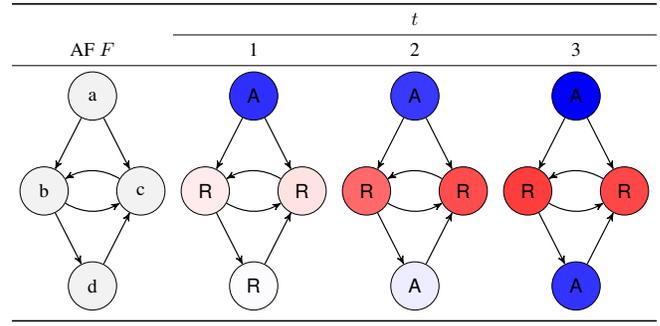

To gain a better understanding of this behaviour we focus on acceptance under the grounded semantics. As shown in Table~\ref{tab:result} AGNN is able to correctly predict the acceptance of all arguments with extremely high confidence. We inspect how arguments in an AF converge over consecutive iterations (as shown in Figure~\ref{fig:message}) and observe three consistent patterns:

\begin{enumerate}
    \item unattacked arguments converge to \emph{accept};
    \item any argument attacked by an argument which is converged to \emph{accept}, converges to \emph{reject};
    \item any argument which is only attacked by arguments which are converged to \emph{reject}, converges to \emph{accept}.
\end{enumerate}

Any argument which is not affected by these procedures converges to \emph{reject} over the course of multiple iterations. Interestingly, each procedural pattern seems to encode some principle of the grounded semantics. Pattern 1 corresponds to the notion that the defence of arguments included in the grounded extension is ‘rooted’ in unattacked arguments \cite{baroni2011introduction}; pattern 2 corresponds with the principle of conflict-freeness; and pattern 3 corresponds with the principle of defence. Since AGNN exhibits these patterns with extremely high confidence predictions (MAE of $7e^{-7}$) on every AF, it seems the model has learned to encode these principles as procedural rules into its message passing algorithm. The procedural rules also correspond with those used in a well-established symbolic labelling algorithm which can be used to find the grounded extension \cite{modgil2009proof}. This algorithm applies the same principles as described in the three observed patterns. It seems AGNN has learned to encode the principles of the grounded semantics and applies these as procedural rules to determine which arguments are contained in the grounded extension.

We also observe that AGNN tries to anticipate the acceptance status of arguments in an opportunistic fashion. Arguments which have not yet converged try to anticipate their status based on information about their neighbourhood. At the first message passing step, only unattacked have enough information to converge. For all other arguments we observe a negative correlation between the degree of incoming attacks and the predicted likelihood of being accepted. Statistically seen, the more incoming attacks an argument has, the higher the chance that it is not defended against one of those attacks\footnote{This encodes the ideas behind ranking-based semantics, in which the numbers of attackers and defenders are used to rank arguments \cite{bonzon2016comparative}.}. It seems that AGNN has learned to infer the amount of incoming attacks from the incoming message and uses this information to anticipate the predicted likelihoods. Additionally, anticipating arguments affect unconverged neighbouring arguments according to the same procedure as mentioned above. An argument attacked by arguments which anticipate \emph{reject} will for instance anticipate \emph{accept} or lower its confidence in anticipating \emph{reject}.

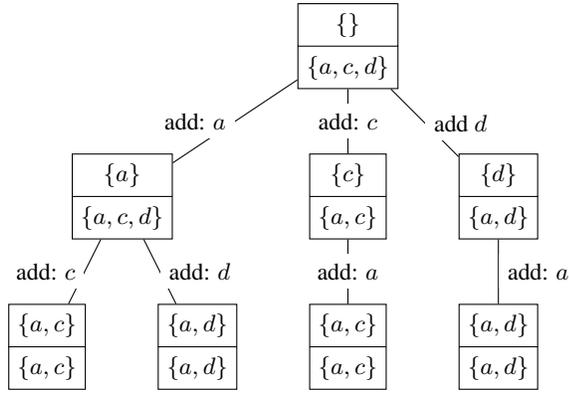
\begin{figure}
 \centering
 \begin{tikzpicture}[level distance=2cm,
  level 1/.style={sibling distance=3cm},
  level 2/.style={sibling distance=2cm},]
  
  \tikzstyle{every node}=[rectangle, fill=white,draw, font=\footnotesize]
  \tikzstyle{data}=[rectangle split,rectangle split parts=2,draw,text centered]
  \node[data] {$\{\}$ \nodepart{second} $\{a,c,d\}$}
    child {node[data] {$\{a\}$ \nodepart{second} $\{a, c,d\}$} 
      child {node[data] {
      $\{a, c\}$ \nodepart{second} $\{a,c\}$} 
  edge from parent node[left,draw=none] {add: $c$}}
      child {node[data] {
      $\{a, d\}$ \nodepart{second} $\{a,d\}$} 
    edge from parent node[right,draw=none] {add: $d$}}
      edge from parent node[left,draw=none] {add: $a$}
    } 
    child {node[data] {
      $\{c\}$ \nodepart{second} $\{a, c\}$} 
  child {node[data] {
      $\{a, c\}$ \nodepart{second} $\{a, c\}$} 
  edge from parent node[draw=none] {add: $a$}}
  edge from parent node[draw=none] {add: $c$}}
    child {node[data, xshift=-10mm] {
      $\{d\}$ \nodepart{second} $\{a, d\}$} 
  child {node[data] {
      $\{a, d\}$ \nodepart{second} $\{a, d\}$} 
  edge from parent node[right,draw=none] {add: $a$}}
  edge from parent node[right,draw=none] {add $d$}};
 \end{tikzpicture}
 \caption{The search tree for enumerating the preferred extensions of the AF $F_{e}$. Each tree node illustrates a set $S$ in the top half and the set of arguments which are constructively accepted w.r.t. $S$ in the bottom half. At each step down the tree a constructively accepted argument is included in $S$ until $S$ becomes an extension (i.e. no constructively accepted argument can extend $S$ any further).}
 \label{fig:treesearch}
\end{figure}

\section{Enumerating Extensions}\label{sec:enumerate}
Motivated by AGNN's ability to predict argument acceptance almost perfectly we expand our scope to the extension enumeration problem. In Section~\ref{sec:message_passing} we showed that AGNN is able to predict the acceptance of arguments under grounded semantics by learning a procedure to enumerate the grounded extension. Since an AF always has one grounded extension, there is a one-to-one mapping between enumerating the extension and deciding acceptance. It seems plausible that AGNN is able to learn a similar procedure under the preferred, stable and complete semantics. However, under these semantics an argument can be contained in multiple extensions and as AGNN can only output a single value per argument there is no straightforward way to directly use AGNN to enumerate all extensions.

To facilitate enumeration under multi-extension semantics we pose enumeration as a search problem and use AGNN to guide a basic search. Starting from an empty set of arguments $S$ we construct a search tree by incrementally adding arguments to $S$ that \emph{extend} $S$  into becoming an extension. When no argument argument can extend $S$ any further we backtrack, select a new argument to extend $S$ and continue the search. Finding extensions with this procedure requires iteratively solving which arguments can extend $S$ into becoming an extension and verifying when $S$ becomes an extension. To address these problems with AGNN we propose the \emph{constructive acceptance} task $\constr_{\sigma}$.

\begin{definition}
\label{def:construct}
Given an AF $F = (A,R)$, a semantics $\sigma$, a set of arguments $S \subseteq A$ and an argument $a \in A$, $a$ is said to be \emph{constructively accepted} w.r.t. $S$ if $S \cup \{a\} \subseteq \bigcup_{\mathcal{E} \in \sigma(F)} \mathcal{E}$

\end{definition}

An argument can only be constructively accepted w.r.t. a set which is subset equal to an extension. Given such a set $S$, an argument $a$ is constructively accepted if it is either contained in $S$ or if adding $a$ to $S$ yields a larger set which is also subset equal to an extension.

\begin{example}
Given the set of arguments $S = \{a\}$ in $F_{e}$, arguments $a$, $c$ and $d$ are constructively accepted w.r.t. $S$ under preferred, complete and stable semantics while only argument $a$ is constructively accepted under grounded semantics.
\end{example}

Consider AF $F=(A,R)$ and semantics $\sigma$. Starting from the empty set $S$ we extend $S$ into an extension by recursively computing which arguments are constructively accepted w.r.t. $S$ and adding one of these arguments to $S$. We use AGNN to approximate the function $f_{\sigma}$ mapping $F$ and $S$ to a binary labelling $f_{\sigma}(F, S)$ indicating for each argument in $A$ if it is constructively accepted w.r.t. $S$. We inform AGNN which arguments are currently in $S$ by initialising the corresponding nodes with a separate embedding $x_{i}$ and we round the computed likelihoods for each argument to a binary answer. 

Each time a constructively accepted argument is added, $S$ is extended into a larger subset of an extension until at some point it becomes equal to an extension. Verifying when $S$ becomes equal to an extension is straightforward under the grounded, preferred and stable semantics. Under these semantics no extension can be a subset of another extension. Therefore $S$ is an extension when all constructively accepted arguments are included in $S$ and no argument can extend it any further. Those extensions are thus found in the leaf nodes of the search tree (as shown in Figure~\ref{fig:treesearch}). Under the complete semantics this principle does not hold for all extensions. Since a complete extension can also be a subset of another complete extension, exhaustively extending $S$ until it becomes an extension will find some, but not all extensions.

Since AGNN provides an approximation, somewhere in the tree search an argument $a$ might falsely be labelled as constructively accepted w.r.t. $S$, yielding the \emph{illegal} set $S \cup \{a\}$. An illegal set is not subset equal to any extension and therefore cannot be extended into a extension. Due to the branching nature of a tree search, a single mislabelled argument early in the search procedure may spawn many illegal sets thereby increasing the risk of an incorrectly enumerated extension. To mitigate this risk while constructing the search tree we stop extending any set $S$ when AGNN's output indicates that $S$ contains a constructively rejected argument (since as long as $S$ is subset equal to an extension all argument in $S$ should be constructively accepted by Definition~\ref{def:construct}). 

\subsection{Experimental Setup and Results}

We alter the training dataset described in Section~\ref{sec:data} to supervise AGNN in learning to predict which arguments are constructively accepted w.r.t. a set of arguments. For each AF we generate a set of arguments $S$ which is subset equal to a randomly selected extension to serve as input feature. The ground-truth labels are determined by taking the union of all extensions which contain $S$ and label each argument as \emph{accepted}. To train AGNN in recognising illegal sets we also generate a set of arguments which is not subset equal to any extension and set the ground-truth label of each argument to \emph{reject}. We train AGNN with the same parameters as described in Section~\ref{sec:training}.

For each AF in the test dataset we use AGNN to construct a search tree and return the sets found in the leaf nodes. Table~\ref{tab:enum} shows AGNN is able to enumerate extensions almost perfectly under most semantics. As anticipated the recall under the complete semantics is relatively low since the search procedure cannot find extensions which are a subset of another extension. However when we include a verification algorithm \cite{besnard2004checking} to enable the identification of such extensions, the recall increases to 0.91. This indicates AGNN has indeed learned the principles of constructing complete extensions but many are not identified as such due to the nature of the search procedure.

\begin{table}
\centering
\setlength\tabcolsep{3pt} 
\begin{tabular}{lrrrr} 

\toprule
 & \multicolumn{4}{c}{$\enum_{\sigma}$} \\
 \cmidrule{2-5}
Metric & \grd & \prf & \stb & \com \\ 
\midrule
Precision & 1.00 & 0.999& 1.00 & 1.00\\
Recall & 1.00 & 0.998 & 0.999 & 0.41 \\

\bottomrule
\end{tabular}
\caption{Extension enumeration results on the test dataset.}
\label{tab:enum}
\end{table}

\section{Discussion}\label{sec:discussion}
\subsection{Related Work}
Existing research on (deep) learning based approaches to argumentation focus mainly on argument mining -- that is, extracting arguments or attacks from natural language text \cite{cocarascu2017identifying} -- instead of solving acceptability problems. The exception is recent work by Kuhlmann and Thimm \shortcite{kuhlmann2019using}, who carried out a feasibility study on the use of a graph convolutional neural network to approximate the credulous acceptance of arguments under the preferred extension. The proposed FM2 model operates as a conventional single forward pass classifier where the number of incoming and outgoing attacks is added to each argument as input features. Thus, FM2 learns to find a correlation between the input features of an argument's neighbourhood and the likelihood of being accepted. In contrast, AGNN learns to perform a sequence of relational inferences which enable it to approximate the acceptance of an argument solely based on the attack structure of an AF. This more general approach greatly outperforms FM2's local focus (see Table~\ref{tab:result}). 

From a machine learning perspective our model is close to Palm et al.~\shortcite{palm2018recurrent} and Gilmer et al.~\shortcite{gilmer2017neural}. Both describe GNNs that learn neural message passing algorithms on problems from symbolic domains. Our model differs since we employ two message functions in order to distinguish between messages sent from attack source to target or vice versa. In addition we show how GNNs can be used to guide a basic search on problems for which multiple solutions exist.

\subsection{Conclusion and Future Work}
We have presented a learning-based approach to determining acceptance of arguments under abstract argumentation semantics, proposing AGNN, which learns a message-passing algorithm to predict the likelihood of an argument being accepted. AGNN can almost perfectly predict the acceptability under different semantics and scales well for larger argumentation frameworks. Furthermore, AGNN can also enumerate all extensions under different semantics very well - for multi-extension semantics, AGNN is used to extend a set of arguments such that it becomes an extension. Analysis of AGNN's behaviour shows that it learns to adhere to basic principles of (ranked) argument semantics as identified in the literature \cite{baroni2011introduction,bonzon2016comparative}, and behaves similarly to a well-known symbolic labelling algorithm for grounded semantics \cite{modgil2009proof}.

Although AGNN does not provide the same theoretical guarantees as a symbolic algorithm, the appeal of a learning-based approach is that it generalises to different problems without needing the expert knowledge of human algorithm designers \cite{li2018combinatorial}. AGNN is a single architecture that can solve different argumentation problems ($\cred$, $\scept$, $\constr$) for different semantics (\grd, \prf, \stb, \com) and on AFs larger than seen during training in constant time, simply by running for more iterations. Additionally, by solving the constructive acceptance problem, AGNN can guide a basic tree search enumerating extensions with a polynomial delay.

For future work, we aim to look at employing AGNN for dynamic argumentation \cite{doutre2018constraints}, looking at whether AGNN can learn, for example, which arguments or attacks should be added or removed to enforce a certain argument's acceptability. 

\clearpage
\bibliographystyle{named}
\bibliography{bib}

\begin{thebibliography}{}

\bibitem[\protect\citeauthoryear{Atkinson \bgroup \em et al.\egroup
  }{2017}]{atkinson2017towards}
K.~Atkinson, P.~Baroni, M.~Giacomin, A.~Hunter, H.~Prakken, C.~Reed, G.~R.
  Simari, M.~Thimm, and S.~Villata.
\newblock Towards artificial argumentation.
\newblock {\em {AI} Magazine}, 38(3):25--36, 2017.

\bibitem[\protect\citeauthoryear{Baroni \bgroup \em et al.\egroup
  }{2011}]{baroni2011introduction}
P.~Baroni, M.~Caminada, and M.~Giacomin.
\newblock An introduction to argumentation semantics.
\newblock {\em Knowledge Engineering Review}, 26(4):365--410, 2011.

\bibitem[\protect\citeauthoryear{Battaglia \bgroup \em et al.\egroup
  }{2018}]{battaglia2018relational}
P.~W. Battaglia, J.~B. Hamrick, V.~Bapst, A.~Sanchez-Gonzalez, V.~Zambaldi,
  M.~Malinowski, A.~Tacchetti, D.~Raposo, A.~Santoro, R.~Faulkner, et~al.
\newblock {Relational inductive biases, deep learning, and graph networks}.
\newblock {\em arXiv e-prints}, arXiv:1806.01261, 2018.

\bibitem[\protect\citeauthoryear{Besnard and
  Doutre}{2004}]{besnard2004checking}
P.~Besnard and S.~Doutre.
\newblock Checking the acceptability of a set of arguments.
\newblock In {\em Proceedings of the 10th International Workshop on
  Non-Monotonic Reasoning (NMR’04)}, pages 59--64, 2004.

\bibitem[\protect\citeauthoryear{Bonzon \bgroup \em et al.\egroup
  }{2016}]{bonzon2016comparative}
E.~Bonzon, J.~Delobelle, S.~Konieczny, and N.~Maudet.
\newblock A comparative study of ranking-based semantics for abstract
  argumentation.
\newblock In {\em Proceedings of the Thirtieth AAAI Conference on Artificial
  Intelligence}, pages 914–--920, 2016.

\bibitem[\protect\citeauthoryear{Charwat \bgroup \em et al.\egroup
  }{2015}]{charwat2015methods}
G.~Charwat, W.~Dvor{\'{a}}k, S.~A. Gaggl, J.~P. Wallner, and S.~Woltran.
\newblock Methods for solving reasoning problems in abstract argumentation -
  {A} survey.
\newblock {\em Artificial Intelligence}, 220:28--63, 2015.

\bibitem[\protect\citeauthoryear{Cocarascu and
  Toni}{2017}]{cocarascu2017identifying}
O.~Cocarascu and F.~Toni.
\newblock Identifying attack and support argumentative relations using deep
  learning.
\newblock In {\em Proceedings of the 2017 Conference on Empirical Methods in
  Natural Language Processing}, pages 1374--1379, 2017.

\bibitem[\protect\citeauthoryear{d'Avila Garcez \bgroup \em et al.\egroup
  }{2015}]{garcez2015neural}
A.~S. d'Avila Garcez, T.~R. Besold, L.~D. Raedt, P.~F{\"{o}}ldi{\'{a}}k,
  P.~Hitzler, T.~Icard, K.~K{\"{u}}hnberger, L.~C. Lamb, R.~Miikkulainen, and
  D.~L. Silver.
\newblock Neural-symbolic learning and reasoning: Contributions and challenges.
\newblock In {\em Proceedings of the 2015 {AAAI} Spring Symposia}, Palo Alto,
  2015. Stanford University.

\bibitem[\protect\citeauthoryear{Doutre and
  Mailly}{2018}]{doutre2018constraints}
S.~Doutre and J.~Mailly.
\newblock Constraints and changes: {A} survey of abstract argumentation
  dynamics.
\newblock {\em Argument {\&} Computation}, 9(3):223--248, 2018.

\bibitem[\protect\citeauthoryear{Dung}{1995}]{dung1995acceptability}
P.~M. Dung.
\newblock On the acceptability of arguments and its fundamental role in
  nonmonotonic reasoning, logic programming and n-person games.
\newblock {\em Artificial Intelligence}, 77(2):321--358, 1995.

\bibitem[\protect\citeauthoryear{Dunne and
  Wooldridge}{2009}]{dunne2009complexity}
P.~E. Dunne and M.~J. Wooldridge.
\newblock Complexity of abstract argumentation.
\newblock In G.~R. Simari and I.~Rahwan, editors, {\em Argumentation in
  Artificial Intelligence}, pages 85--104. Springer, 2009.

\bibitem[\protect\citeauthoryear{Gaggl \bgroup \em et al.\egroup
  }{2020}]{gaggl2020design}
S.~A. Gaggl, T.~Linsbichler, M.~Maratea, and S.~Woltran.
\newblock Design and results of the second international competition on
  computational models of argumentation.
\newblock {\em Artificial Intelligence}, 279:103193, 2020.

\bibitem[\protect\citeauthoryear{Gilmer \bgroup \em et al.\egroup
  }{2017}]{gilmer2017neural}
J.~Gilmer, S.~S. Schoenholz, P.~F. Riley, O.~Vinyals, and G.~E. Dahl.
\newblock Neural message passing for quantum chemistry.
\newblock In {\em Proceedings of the 34th International Conference on Machine
  Learning}, volume~70, pages 1263--1272, 2017.

\bibitem[\protect\citeauthoryear{Hochreiter and
  Schmidhuber}{1997}]{hochreiter1997lstm}
S.~Hochreiter and J.~Schmidhuber.
\newblock Long short-term memory.
\newblock {\em Neural Computation}, 9(8):1735--1780, 1997.

\bibitem[\protect\citeauthoryear{Kipf and Welling}{2017}]{kipf2017semi}
T.~N. Kipf and M.~Welling.
\newblock Semi-supervised classification with graph convolutional networks.
\newblock In {\em 5th International Conference on Learning Representations)},
  2017.

\bibitem[\protect\citeauthoryear{Kuhlmann and Thimm}{2019}]{kuhlmann2019using}
I.~Kuhlmann and M.~Thimm.
\newblock Using graph convolutional networks for approximate reasoning with
  abstract argumentation frameworks: A feasibility study.
\newblock In {\em Proceedings of the 13th international conference on Scalable
  Uncertainty Management (SUM)}, pages 24--37, 2019.

\bibitem[\protect\citeauthoryear{Li \bgroup \em et al.\egroup
  }{2018}]{li2018combinatorial}
Z.~Li, Q.~Chen, and V.~Koltun.
\newblock Combinatorial optimization with graph convolutional networks and
  guided tree search.
\newblock In {\em Proceedings of the 32nd International Conference on Neural
  Information Processing Systems}, pages 537--546, 2018.

\bibitem[\protect\citeauthoryear{Loshchilov and
  Hutter}{2019}]{loshchilov2017decoupled}
I.~Loshchilov and F.~Hutter.
\newblock Decoupled weight decay regularization.
\newblock In {\em 7th International Conference on Learning Representations)},
  2019.

\bibitem[\protect\citeauthoryear{McKay and Piperno}{2014}]{mckay2014practical}
B.~D. McKay and A.~Piperno.
\newblock Practical graph isomorphism, {II}.
\newblock {\em Journal of Symbolic Computation}, 60:94--112, 2014.

\bibitem[\protect\citeauthoryear{Modgil and Caminada}{2009}]{modgil2009proof}
S.~Modgil and M.~Caminada.
\newblock Proof theories and algorithms for abstract argumentation frameworks.
\newblock In I.~Rahwan and G.~R. Simari, editors, {\em Argumentation in
  Artificial Intelligence}, pages 105--129. Springer, 2009.

\bibitem[\protect\citeauthoryear{Niskanen and
  J{\"a}rvisalo}{2019}]{niskanen2019mu}
A.~Niskanen and M.~J{\"a}rvisalo.
\newblock $\mu$-toksia: Sat-based solver for static and dynamic argumentation
  frameworks.
\newblock \url{https://bitbucket.org/andreasniskanen/mu-toksia}, 2019.

\bibitem[\protect\citeauthoryear{Palm \bgroup \em et al.\egroup
  }{2018}]{palm2018recurrent}
R.~B. Palm, U.~Paquet, and O.~Winther.
\newblock Recurrent relational networks.
\newblock In {\em Proceedings of the 32nd International Conference on Neural
  Information Processing Systems}, pages 3372--3382, 2018.

\bibitem[\protect\citeauthoryear{Pascanu \bgroup \em et al.\egroup
  }{2013}]{pascanu2013difficulty}
R.~Pascanu, T.~Mikolov, and Y.~Bengio.
\newblock On the difficulty of training recurrent neural networks.
\newblock In {\em Proceedings of the 30th International Conference on Machine
  Learning ({ICML})}, pages 1310--1318, 2013.

\bibitem[\protect\citeauthoryear{Powers}{2011}]{powers2011evaluation}
D.~Powers.
\newblock Evaluation: From precision, recall and f-measure to roc,
  informedness, markedness \& correlation.
\newblock {\em Journal of Machice Learning Technology}, 2:2229--3981, 01 2011.

\bibitem[\protect\citeauthoryear{Selsam \bgroup \em et al.\egroup
  }{2019}]{selsam2018learning}
D.~Selsam, M.~Lamm, B.~B{\"{u}}nz, P.~Liang, L.~de~Moura, and D.~L. Dill.
\newblock Learning a {SAT} solver from single-bit supervision.
\newblock In {\em 7th International Conference on Learning Representations
  ({ICLR})}, 2019.

\bibitem[\protect\citeauthoryear{Smith}{2017}]{smith2017cyclical}
L.~N. Smith.
\newblock Cyclical learning rates for training neural networks.
\newblock {\em IEEE Winter Conference on Applications of Computer Vision
  (WACV)}, 2017.

\end{thebibliography}

\end{document}